\pdfoutput=1

\documentclass[11pt]{article}

\usepackage[preprint]{acl}

\usepackage{times}
\usepackage{latexsym}
\usepackage{tabularray}
\usepackage{amssymb}
\usepackage{pifont}
\usepackage[T1]{fontenc}

\usepackage[utf8]{inputenc}

\usepackage{microtype}

\usepackage{inconsolata}

\usepackage{graphicx}
\usepackage{multirow}
\usepackage{listings}

\title{BRIT: Bidirectional Retrieval over Unified Image-Text Graph}

\author{
 \textbf{Ainulla Khan\thanks{These authors contributed equally to this work.}}\hspace{0.5cm}
 \textbf{Moyuru Yamada\textsuperscript{*}}\hspace{0.5cm}
 \textbf{Srinidhi Akella\thanks{The author contributed while at Fujitsu Research of India.}}
 \\
\\
 Fujitsu Research India
 \\
 \{ainulla.khan, yamada.moyuru\}@fujitsu.com 
 }

\begin{document}
\maketitle
\begin{abstract}

Retrieval-Augmented Generation (RAG) has emerged as a promising technique to enhance the quality and relevance of responses generated by large language models. While recent advancements have mainly focused on improving RAG for text-based queries, RAG on multi-modal documents containing both texts and images has not been fully explored. Especially when fine-tuning does not work. This paper proposes \textit{BRIT}, a novel multi-modal RAG framework that effectively unifies various text-image connections in the document into a multi-modal graph and retrieves the texts and images as a query-specific sub-graph. By traversing both image-to-text and text-to-image paths in the graph, BRIT retrieve not only directly query-relevant images and texts but also further relevant contents to answering complex cross-modal multi-hop questions. To evaluate the effectiveness of BRIT, we introduce MM-RAG\footnote{MM-RAG: \href{https://ast-fri.github.io/BRIT/}{https://ast-fri.github.io/BRIT/}} test set specifically designed for multi-modal question answering tasks that require to understand the text-image relations. Our comprehensive experiments demonstrate the superiority of BRIT, highlighting its ability to handle cross-modal questions on the multi-modal documents.

\end{abstract}

\section{Introduction}


\begin{figure*}[t]
    \centering
    \includegraphics[width=1.0\textwidth]{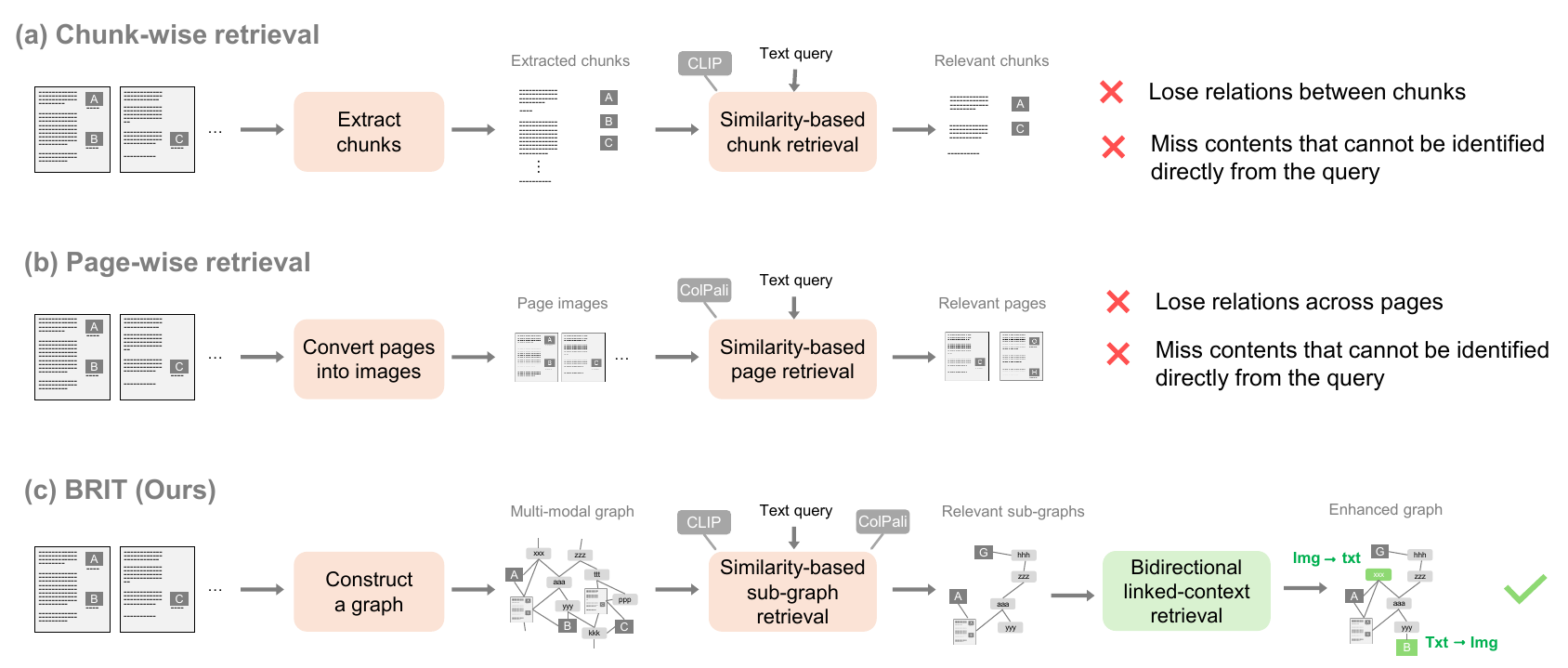}
    \caption{Comparison between standard multi-modal retrieval methods and our BRIT. The standard methods simply retrieve relevant chunks or pages based on the similarity in the embedding space. These methods may struggle with complex questions that require understanding the connections between texts and images across multiple pages. BRIT retrieves not only directly relevant content with similarity (any embedding model, such as CLIP or ColPali, can be used), but also indirectly connected information essential for answering the query by traversing image-to-text and text-to-image links bidirectionally.}
    \label{fig:comaprison}
\end{figure*}

Retrieval-Augmented Generation (RAG) \cite{rag} has emerged as a promising approach for enhancing Large Language Models (LLMs) by grounding their responses in external knowledge.  This technique retrieves relevant information from external sources, such as proprietary company documents, and provides it to the LLM as context for generating more informed and accurate responses.




While recent efforts \cite{golden-retriever, adaptive-rag,self-rag,c-rag} have largely focused on improving textual RAG, effectively incorporating visual information remains a challenge. This capability is crucial for comprehensive understanding of documents like company brochures, websites, and presentations, where visual content plays a vital role in conveying information.

Multi-Modal RAG (MM-RAG) aims to address this challenge by retrieving both relevant texts and images for response generation. However, MM-RAG faces difficulties in effectively aligning visual content with textual queries. A common approach (e.g., Fig. \ref{fig:comaprison} (a)) relies on embedding-based similarity \cite{openclip} between the query and the images. This approach often fails in enterprise settings where queries contain company-specific terms (e.g., product or person names) that are absent in the training data of pre-trained embedding models. For example, answering the question "\textit{Does the codename BRIT have a logo on the top?}" requires retrieving an image of BRIT. However, standard embedding models may not effectively associate the codename with its corresponding image based solely on similarity.  Recent works \cite{colpali,m3docrag} have explored retrieving relevant pages by treating each page as an image and computing its similarity to the query  (Fig. \ref{fig:comaprison} (b)). This page-wise retrieval struggles when relevant contents are spanned across multiple pages, as it relies on the similarity between the entire page and the query.
Furthermore, these approaches lose crucial text-image associations when the retrieved information is provided as input.


The existing methods \cite{mu-rag,mm-mh-qa,uni-rag} often rely on training or fine-tuning techniques to address these limitations, but these approaches may not be effective in the enterprise settings, particularly when they are frequently updated.
Thus, other approaches to connect images to their textual descriptions must be explored and evaluated.

Graph RAG \cite{hippo-rag} has been proposed to overcome the shortcomings of the standard RAG. Recent studies \cite{holmes,graph-rag} have demonstrated its superior accuracy in textual domains. Graph representations provide a natural framework for modeling diverse relationships between textual and visual contents; however, The effectiveness of graph-based methods in multi-modal settings remains underexplored. A key challenge is how to effectively traverse modalities to reach the information needed to answer a question.

In this paper, we propose \textit{BRIT} (\textbf{B}idirectional \textbf{R}etrieval over Unified \textbf{I}mage-\textbf{T}ext Graph), a novel multi-modal RAG framework illustrated in Fig. \ref{fig:comaprison} (c). BRIT constructs a multi-modal graph from document text and images, integrating diverse text-image relationships.  Given an input query, BRIT extracts a relevant sub-graph based on node-query and edge-query similarities.  Crucially, by bidirectionally traversing image-to-text and text-to-image links, BRIT expands this initial sub-graph, retrieving not only directly relevant content, but also indirectly connected information essential for answering the query.
Unlike many existing methods, our approach does not require training or fine-tuning of either the LLM or the retriever.
This paper presents a comprehensive evaluation of the effectiveness of different text-image linking strategies and their combinations on documents containing both text and images. To asses the performance of MM-RAG in the enterprise settings, we construct MM-RAG test set, a new benchmark comprising 500 complex questions that necessitate cross-modal, multi-hop retrieval to identify the key information. 


Our main contributions can be summarized as follows:
\begin{itemize}
    \item We propose BRIT, a novel multi-modal RAG framework integrates diverse text-image links within a unified graph. This enables effective retrieval of relevant texts and images for answering complex cross-modal questions.
    \item We construct MM-RAG test set, a new test set to assess complex questions that require understanding both texts and images and their connections.
    \item We conduct a comprehensive evaluation of multi-modal graph RAG, analyzing the impact of different text-image linkings on the MM-RAG test set.
\end{itemize}

\section{Related Work}

\begin{figure*}[t]
    \centering
    \includegraphics[width=1\textwidth]{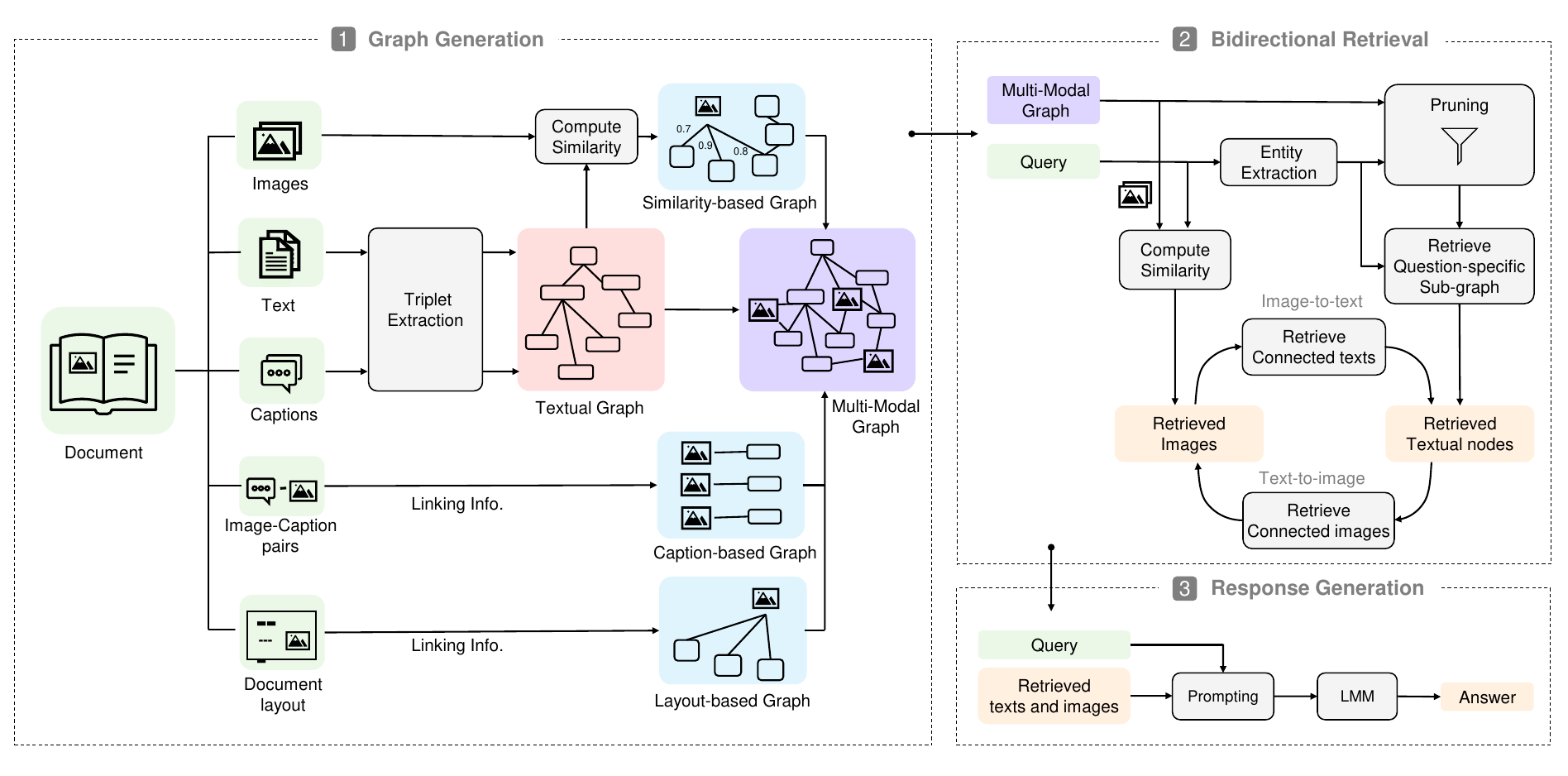}
    \caption{A overview of our Multi-Modal Graph RAG, BRIT consisting of 1) Graph Generation, 2) Bidirectional Retrieval, and 3) Response Generation. We first construct a multi-modal graph based on extracted textual triplets and connections between texts and images in various aspects. Then, query-relevant triplets and images are retrieved with a given query. Finally, the retrieved contexts are fed into LMM with our prompting.}
    \label{fig:framework}
\end{figure*}

\subsection{Multi-Modal RAG}
Recent methods mostly focus on effective techniques for training or fine-tuning a multi-modal encoder and retriever to improve the retrieval performance. \citet{mu-rag} introduces MuRAG which has a fine-tuned multi-modal encoder to convert images and texts into a sequence of vectors. These vectors are fed to a decoder for text generation. UniRAG \cite{uni-rag} also uses fine-tuned universal retriever to retrieve the images with a text query. They are trained on a large dataset and evaluated on it. This approach may not work for real applications which need to handle texts and corresponding but uncorrelated images (e.g., product names and their images) since it is difficult to update the retriever frequently. Standard Multi-Modal RAG pipeline commonly employs the CLIP \cite{openclip} as a retriever and encodes text chunks and images extracted from a document separately to construct a multi-modal vector DB. This approach cannot consider the relations between the texts and images during retrieval.


\subsection{Graph RAG}
Some prior works \cite{hippo-rag, holmes,g-retriever} use knowledge graph for RAG. The knowledge graph allows them to retrieve a query-specific textual entities, improving the QA performance. \citet{graph-rag} introduces Graph RAG for the query-focused summarization task. They construct a textual graph from a document by extracting triplets using a LLM and retrieve the sub-graph from the entire graph with the query. The retrieved triplets are converted to the texts and fed into a LLM for reasoning. While the textual graph can be naturally extended to a multi-modal graph, the integration of various text-image connections and query-specific multi-modal graph retrieval have not been explored enough.

\subsection{Multi-Modal Graph and LMMs}
Several recent works \cite{mm-mh-qa, mm-graph-sum} attempt to use multi-modal graph for summarization task and QA task with Large Multi-modal Model (LMM). They train their neural networks on the specific datasets and test their trained models on them. Unlike them, we do not train or fine-tune any neural networks. We instead focus on comprehensively evaluating the effectiveness of various text-image links and their combinations for RAG on a new test set we built to asses them with complex questions on multi-modal documents containing texts and images.


\section{BRIT: Multi-Modal Graph RAG} 

Our framework, BRIT, enables multi-modal retrieval by integrating images and related texts using graphs. The process consists of three key steps: graph generation, bidirectional retrieval, and response generation, as shown in Fig. \ref{fig:framework}. First, a textual graph is constructed from document texts and images are linked with their corresponding textual nodes, forming a unified multi-modal graph. Next, a query-relevant sub-graph is retrieved through the formulation of Prize-Collecting Steiner Tree (PCST) optimization problem. Subsequently, we expand this initial sub-graph by traversing from query-relevant textual nodes to connected images or from query-relevant images to the connected textual nodes to retrieve the cross-modal context. Finally, the retrieved multi-modal context with the input query is fed into an LMM to generate the final response.


\subsection{Graph Generation}

For a given text, we extract triplets consisting of <subject, relation, object>. The subject and object are named entities (e.g., person names, locations, dates). From these extracted triplets, we construct a set of disjoint textual graphs $G_{d} = \{g_{1}, .., g_{n}\}$. For each graph in $G_{d}$ we link textual nodes to their relevant images.


We consider the following three methods for linking textual nodes and images: 

\noindent \textbf{Captions-based (CA)}: For each image-caption pair, we generate textual nodes by extracting named entities from the captions using LLMs. The image is then linked to these generated textual nodes.

\noindent \textbf{Similarity-based (SI)}: We use the CLIP encoder \cite{openclip} to embed the attributes of both the textual nodes and images. The images are linked to the textual nodes based on their cosine similarity scores.

\noindent \textbf{Layout-based (LA)}: Textual nodes are linked to images based on the page or section layout. In page-based linking (LP), images are connected to textual nodes if both the image and the text content of the textual nodes are on the same page. In section-based linking (LS), the image is linked to textual nodes if they appear within the same section of the document.

Note that other linking methods can also be integrated. For example, each page can be converted into an image, as in page-wise retrieval, and then linked to the corresponding text nodes on that page.

\begin{figure*}[t]
    \centering
    \includegraphics[width=1.0\textwidth]{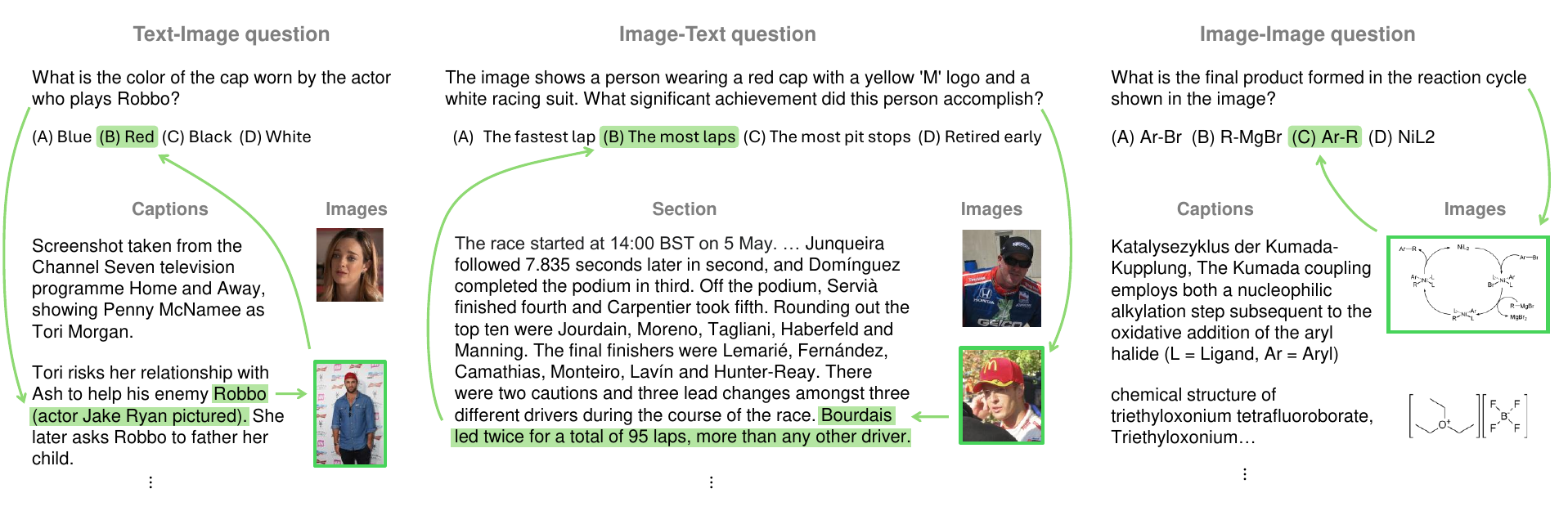}
    \caption{Examples of generated questions in MM-RAG test set. This test set contains 3 types of questions, 1) Text-Image questions, 2) Image-Text questions, and 3) Image-Image questions. Text-Image and Image-Text questions require an image or text for an answer, however the specific image or text cannot be directly identified from the question. First, it is necessary to identify the text or image related to the question, and then to find the image or text linked to that. Green arrows show an expected path to reach the answer.}
    \label{fig:questions}
\end{figure*}

\subsection{Bidirectional Retrieval}

Our retrieval process consists of two steps: query-relevant sub-graph retrieval and linked-context retrieval. Given a query, first a sub-graph is retrieved and then a bidirectional retrieval process follows based on two pathways over the generated multi-modal graph: (1) Text-to-image retrieval, and (2) Image-to-text retrieval. 

\noindent \textbf{Sub-graph retrieval}.
Given $G_{d}$, to retrieve query-specific sub-graphs $G_{q}$ from $G_{d}$, we first extract named entities from the query using a 1-shot LLM prompt. Then, the entities of query and the textual nodes of $G_{d}$ are embedded using an embedding model (e.g., CLIP and BLIP). Next, we prune $G_{d}$ to obtain $G_{q} = \{g_{q_{1}}, .., g_{q_{n}}\}$. Given a disjoint graph $g_{m}$, the pruning process begins by computing cosine similarity scores between the entity embeddings of the query and the nodes of $g_{m}$, then among these similarity scores we compute the highest similarity score and if the score exceeds a pre-defined threshold then the graph  $g_{m}$ is considered query-relevant and retained as a sub-graph $g_{q_{m}}$. This pruning process is repeated for all disjoint graphs.
Subsequently, we apply the PCST optimization \cite{pcst} to filter out irrelevant entities while preserving the overall structure. Finally, we consider the nodes of the refined query-relevant sub-graphs as retrieved textual nodes.

Among various graph-retrieval approaches we employ a PCST-based approach for graph retrieval because it takes into account the edge attributes, helping with image retrieval. The PCST algorithm retrieves a sub-graph that is relevant to the query, even if some nodes have a lower similarity score with the query. This improves the retrieval of connected images to those specific nodes.

\noindent \textbf{Linked-context retrieval}. This is a crucial step to retrieve not only directly relevant content, but also indirectly connected information essential for answering the query.

\noindent (1) Text-to-image retrieval: Using the retrieved textual nodes, we perform a similarity score-based traversal for image retrieval. Among multiple images we select the image whose connected textual node is most similar to the query. 

\noindent (2) Image-to-text retrieval: We compute similarity between the query and images in the embedding space, selecting top-k images as relevant images based on the similarity scores which is followed by the retrieval of connected textual nodes.

BRIT can be easily extend to argentic framework which iteratively retrieves the linked-content.

\subsection{Response Generation}
We convert the retrieved triplets into sentences, as LLMs are trained to process natural language, by concatenating the triplet's source node entity, the relation, and the destination node entity. The resulting sentences, separated by a delimiter, along with the retrieved images and their relevant texts, are then included in an input prompt, which is fed into LMM to get a response.

\section{MM-RAG Test Set}
\label{sec:testset}


\begin{figure*}[t]
    \centering
    \includegraphics[width=1.0\textwidth]{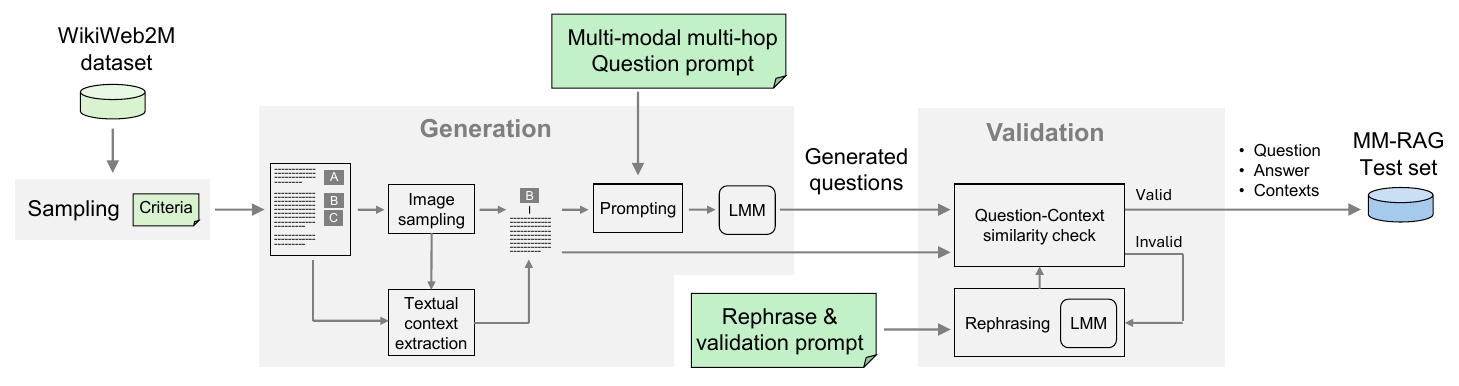}
    \caption{Question generation pipeline for MM-RAG test set. First, 100 Wikipedia pages are carefully sampled from WikiWeb2M dataset. Then, an image is selected and fed into a LMM along with the relevant text to generate a question with a specific prompt we designed. We also validate and refine the generated questions with LMM.}
    \label{fig:ques_gen}
\end{figure*}

This section details MM-RAG test set, a new test set we have constructed, and the methodology employed for its construction. MM-RAG test set is constructed to evaluate RAG on multi-modal documents containing texts and images with complex cross-modal questions, which require to understand connections between texts and images. Some recent works have proposed multi-modal question answering datasets \cite{webqa,mm-mh-qa} which require multi-hop reasoning over different modalities, however their questions are not explicitly designed to require traversing different modalities. 

Our MM-RAG test set contains three types of questions as shown in Fig. \ref{fig:questions}. 1) Text-Image questions are the question which requires a relevant image to answering but that image cannot be directly identified by the question, whereas 2) Image-Text questions require the relevant texts to answering but texts cannot be directly identified by the question. These Text-Image and Image-Text questions are challenging cross-modal multi-hop questions because they require to understand not only texts and images but also their relationships to connect a reasoning path. 3) Image-Image questions are simple question that serves as a baseline and can be matched with an image and answered based only on that image. 

\subsection{Question Generation}

We design a question generation pipeline as shown in Fig. \ref{fig:ques_gen}. We employ WikiWeb2M dataset \cite{wikiweb2m} as a source data. WikiWeb2M dataset is built for multi-modal content understanding tasks with many-to-many text and image relationships. It includes the page title, section titles, section text, images and their captions, and so on. We first sample 100 Wikipedia pages that contains at least 3 images and 1 section from a validation split. In more detail, average number of images and sections per sample is 5.61 and 8.83. Then, an image is randomly picked for each question type (3 images in total in each sample). The picked image is fed into a LMM (Large Multi-Modal Model) along with the relevant texts with a specific prompt we designed to generate a question and an answer pair. Specifically, we use GPT-4o (gpt-4o-2024-05-13) to handle both texts and images. 

For both Text-Image and Image-Text questions, we used image captions and the text within the same section as the image as sources for question generation.  Therefore, each question type (Text-Image and Image-Text) has a total of 200 questions: 100 generated from image captions and 100 generated from text within the same section as the image.
Finally, we have generated 500 questions in total. To evaluate the retrieval performance (not QA accuracy), we have also stored an image and relevant texts used to generate the questions.
Although the WikiWeb2M dataset does not have a concept of page, to test the page-level linking we divided a single Wiki page into multiple pages by setting the number of triplets each page can accommodate. 


\subsection{Validation}
We designed a validation process to ensure the validity of the generated questions and answers. Our questions must be designed so that the evidence (text or image) required to answer them cannot be directly identified from the question alone. Therefore, we calculated the similarity between the generated question and the source text used for its generation. We iteratively revised and validated the question using a LMM until this similarity fell below a predefined threshold.

\subsection{Named Entity Anonymization}
A significant challenge in evaluating RAG performance with pre-trained retrievers lies in the uncertain knowledge coverage of the retriever, stemming from a lack of control over its training data. To simulate an enterprise setting where queries often contain company-specific terminology, we generated additional test questions using a technique we call Named Entity Anonymization (NEA).  The process begins by creating a simple image-image question and identifying a phrase within that question that targets a specific image. This phrase is then replaced with a imaginary name, not found in real-world data. Finally, a relationship between the imaginary name and the original phrase is added to the document. We generated 100 question pairs (200 in total); each pair consists of one question with NEA and a corresponding question without NEA.

\begin{table*}[t]
  \centering
    \begin{tabular}{l|c|ccc}
    \hline
    \multirow{3}{*}{Methods} & Simple QA & \multicolumn{3}{c}{Complex cross-modal multi-hop QA}  \\ \cline{2-5}
     & Image-Image & Text-Image & Image-Text & Average \\ \hline
    CLIP & 0.81 & 0.70 & 0.76 & 0.73  \\
    BLIP2 & \textbf{0.82} & 0.68 & 0.78 & 0.73  \\ \hline
    BRIT (Ours-Full) & 0.78 & 0.72 & \textbf{0.89} & 0.80  \\
    BRIT (w/o CA) & 0.79 & \textbf{0.74} & 0.79 & 0.76 \\
    BRIT (w/o LS) & 0.78 & 0.71 & 0.82 & 0.76 \\
    BRIT (w/o SI) & 0.78 & 0.73 & \textbf{0.89} & \textbf{0.81} \\
    BRIT (No-linking) & 0.81 & 0.66 & 0.56 & 0.61 \\
    \hline
    \end{tabular}
  \caption{\label{citation-guide}
    Question answering accuracy on MM-RAG test set. CA: Caption-based, LS: Layout-based (section), and SI: Similarity-based. Full denotes CA+LS+SI.
  }
  \label{tbl:qa}
\end{table*}

 \begin{table*}[t]
  \centering
    \begin{tabular}{l|cccc|c|c|c|cc}
    \hline
    \multirow{2}{*}{Methods} & \multicolumn{4}{c|}{Text-Image linking} & \multicolumn{3}{c|}{Recall ratio} & \multicolumn{2}{c}{Retrieved} \\ \cline{2-10}
        & CA & LP & LS & SI & Text-Image & Image-Text & Overall & Words & Images \\ \hline
    CLIP ($k$=2) & - & - & - & - & 0.81 & 0.37 & 0.59 & 206.2 & 2.00  \\
    BLIP2 ($k$=2) & - & - & - & - & 0.75 & 0.46 & 0.60 & 257.5 & 2.00  \\ \hline
     &  &  &  &  &  0.61 & 0.02 & 0.32 & 18.9 & 1.00 \\
                          & \checkmark &  &  & & 0.76 & 0.44 & 0.60 & 101.2 & 1.43 \\
                          &  & \checkmark &  & & 0.79 & 0.30 & 0.54 & 262.5 & 1.80 \\
                          & &  & \checkmark  & & 0.82 & 0.35 & 0.58 & 195.7 & 1.69 \\
                          &  &  &  & \checkmark & 0.70 & 0.09 & 0.39 & 55.4 & 1.35 \\ \cline{2-10}
                          BRIT & \checkmark & \checkmark &  & & 0.84 & 0.67 & 0.76 & 303.0 & 1.82 \\
                          (Ours)& \checkmark &  & \checkmark & & 0.87 & 0.71 & 0.79 & 245.7 & 1.78 \\
                          & \checkmark &  &  & \checkmark & 0.72 & 0.49 & 0.64 & 277.1 & 1.55 \\ \cline{2-10}
                          & & \checkmark &  & \checkmark & 0.82 & 0.31 & 0.56 & 213.1 & 1.85 \\
                          & &  & \checkmark & \checkmark & 0.87 & 0.35 & 0.59 & 119.7 & 1.75 \\ \cline{2-10}
                          & \checkmark & \checkmark &  & \checkmark & 0.86 & 0.68 & 0.77 & 316.0 & 1.86 \\
                              & \checkmark &  & \checkmark & \checkmark & \textbf{0.88} & \textbf{0.72} & \textbf{0.80} & 261.0 & 1.83 \\
    \hline
    \end{tabular}
  \caption{\label{citation-guide}
    Recall ratio in Retrieval. We evaluate various text-image linking and their combinations in terms of the recall rate and the number of retrieved words and images. CA: Caption-based, LP: Layout-based (page), LS: Layout-based (section), and SI: Similarity-based. Caption and Section denotes that question generated from a caption and texts in a section. 
    The baseline retrieves top-$k$ images based on the similarity.
  }
  \label{tbl:recall}
\end{table*}

\section{Experimental Settings}
\subsection{Benchmark}
We use MM-RAG test set described in Sec.\ref{sec:testset} to evaluate RAGs on multi-modal documents containing texts and images. MM-RAG test set is a new test set we built based on WikiWeb2M \cite{wikiweb2m} and contains 500 questions for 100 Wikipedia page samples. There are 3 types of questions, Text-Image questions, Image-Text questions, and Image-Image questions. MM-RAG covers a wide variety of topics and images, containing not only natural images but also drawings and diagrams. This new test set allows us to evaluate RAG on various settings. 

\subsection{Baseline}
We utilize two multi-modal baselines using the embeddings from CLIP and BLIP2 \cite{blip2}, where retrieval is purely based on similarity scores, treating text and images independently. For the text, we divide it into chunks and generate embeddings for each chunk using the multi-modal encoders. Similarly, the images are also encoded using the same enocders. These text and image embeddings are then used for query-based similarity retrieval. 
For each baseline, the retrieved texts with top-2 relevant images are then fed into an LMM to generate the final response.
Since our test set does not include PDF or image-formatted pages, we evaluate page-based and section-based linking instead of methods that convert each page into an image (e.g., ColPali \cite{colpali}).


\subsection{Implementation Details}
We use GPT-4o for triplet extraction. We also employ OpenCLIP \cite{openclip} with ViT-H/14 trained on LAION-2B \cite{laion} as our vision-and-language model to compute the similarity between the text and the image. We link every image with top 3 textual nodes as the similarity-based graph.
Under retrieval, we set the threshold of 0.75 to prune $G_{d}$ and while refining $G_{q}$ via PCST, we set $k=5$ for selecting top $k$ nodes and edges with the edge cost $C_{e}$ set to 0.5. For retrieving images based on query-image similarity we select the top-1 image.
We use Gemini 1.5 Flash with temperature of 0 as our LMM for inference.


\subsection{Evaluation Metrics}
We report recall ratio and QA accuracy on our MM-RAG test set. 
\section{Results and Analysis}


\begin{figure*}[!h]
    \centering
    \includegraphics[width=1.0\textwidth]{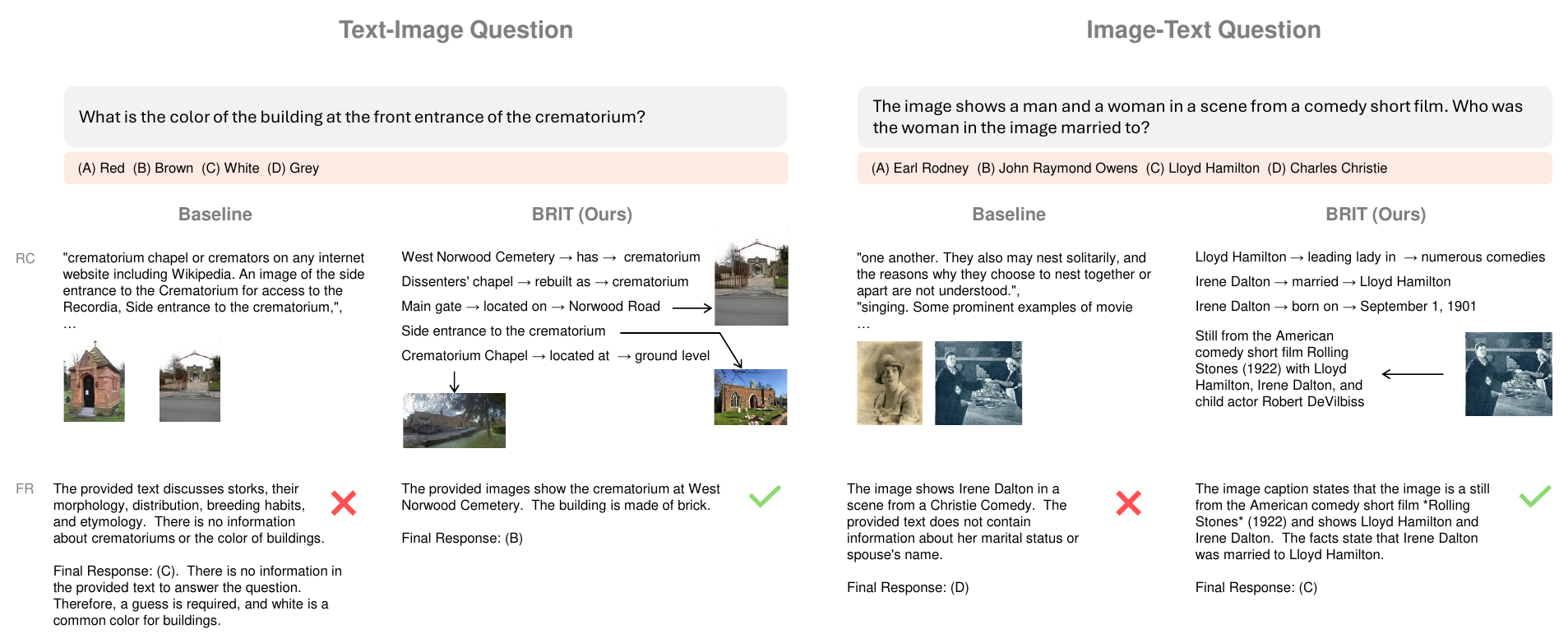}
    \caption{Examples of results on Text-Image and Image-Text questions. BRIT reaches an answer by traversing the multi-modal connections. RC and FR denotes the retrieved contexts and the final response.}
    \label{fig:examples}
\end{figure*}

\subsection{QA and Recall Performances}

We first show question answering performance of all methods in Table \ref{tbl:qa} and then discuss on recall performance shown in Table \ref{tbl:recall}.

BRIT outperforms the baselines in overall QA accuracy on complex question (0.73 vs 0.81), as shown in Table \ref{tbl:qa}.  Image-Image questions are the simple question which does not require the cross-modal multi-hop retrieval. Thus, the baseline methods with the simple retrieval of the top-2 most relevant images achieves the highest accuracy, while BRIT only retrieve 1 image with the similarity. 
However, the baseline struggles with Image-Text questions, resulting in a large drop in performance compared to other cases. Table \ref{tbl:qa} also shows performance gains from caption-based and layout-based linking, but no gain from similarity-based linking.

Recent LMMs can answer the complex multi-modal questions when we give enough contexts even if redundant texts and images are included in the contexts. Thus, we investigate the recall rates of various methods as shown in Table \ref{tbl:recall}. 
BRIT demonstrates superior performance on both Text-Image and Image-Text questions compared to the baselines, achieving a significant overall improvement (0.6 vs. 0.8). Similar to the QA accuracy shown in Table \ref{tbl:qa} the combination of the caption-based and structure-based linking shows a large parformance gain.
Furthermore, Table \ref{tbl:recall} reveals that section-based linking consistently outperforms page-based linking, highlighting the importance of inter-page connections.

\subsection{Recall Performance with NEA}
To simulate a real-world enterprise use case where we need to retrieve a product image with its name, we employ Named Entity Anonymization (NEA) to anonymize the phrase identifying the target image in simple image-image questions. 
Table \ref{tbl:nea} demonstrates that baseline accuracies are significantly reduced with NEA, as these methods rely solely on similarity-based image retrieval. 
In contrast, BRIT exhibits superior performance to the baselines in both scenarios (with and without NEA). Notably, since BRIT establishes connections between images and their corresponding named entities within the document, its performance even improves with NEA.

\begin{table}[t]
  \centering
    \begin{tabular}{l|cc}
    \hline
    \multirow{3}{*}{Methods} & \multicolumn{2}{c}{Recall ratio}  \\ \cline{2-3}
     & w/o NEA & w/ NEA \\ \hline
    CLIP ($k$=2) & 0.82 & 0.58 \\
    BLIP2 ($k$=2) & 0.84 & 0.61 \\ \hline
    BRIT (Ours-Full) & \textbf{0.88} & \textbf{0.98}  \\
    \hline
    \end{tabular}
  \caption{\label{citation-guide}
    Recall ratio for questions without and with NEA (Named Entity Anonymization). Full denotes CA+LS+SI.
  }
  \label{tbl:nea}
\end{table}

\subsection{Qualitative Evaluations}
Figure. \ref{fig:examples} shows some examples of the results on Text-Image and Image-Text questions.
On Text-Image questions (left), the baseline retrieves similar images matched with the query, however a correct image cannot be retrieved. BRIT can find the query-relevant texts and then retrieve the connected images to reach the answer. On Image-Text questions (right), both the baseline and BRIT retrieve the relevant image, however the baseline cannot find the relevant texts because an important keyword 'Irene Dalton' is not in the question. Our BRIT finds the relevant image first, then discovers relevant texts by following the link between the image and the text.



\section{Conclusion}
We have proposed a novel multi-modal RAG framework, \textit{BRIT} which unifies various text-image connections into a multi-modal graph and retrieves the texts and images as a query-specific sub-graph from the multi-modal graph. 
Unlike the standard multi-modal RAG which separately retrieves texts and images with the similarity, BRIT retrieve not only directly query-relevant images and texts but also discover further relevant contents by traversing both image-to-text and text-to-image links extracted from a document bidirectionally.
This paper has comprehensively evaluated the effectiveness of the various links and their combinations for RAG on multi-modal document using a new test set, MM-RAG test set we built, which contains complex cross-modal multi-hop questions, requiring to understand the text-image relations. We also demonstrated a significant decrease in recall performance of existing methods when using questions containing unknown terms.


\section{Limitations}

The test dataset we used is based on the WikiWeb2M dataset, which wasn’t originally created for retrieval tasks involving specific person, product, or company names. However, RAG is meant to handle question-answering tasks on documents that are often focused on a specific domain. To properly evaluate our method, we need to broaden the scope by testing other multi-modal linking approaches. For example, when working with specific document manuals containing only images a scene graph could be helpful for extracting objects in image, then generate object-relevant texts and merge with the original image. 
Text-Image questions and Image-Text questions are designed such that the question requires a relevant image/text to answering but that image/text cannot be directly identified by the question. However, due to the nature of the Wikipedia page if only few images are in the same Wikipedia page and they are different, the target image can be identified by the some key words in the question. We will work on more challenging settings.

\newpage
\clearpage

\bibliography{custom}

\newpage
\clearpage
\appendix

\section{Prompt for Question Generation}
\label{sec:appendix}

We show a prompt we designed to generate three types of questions, 1) Text-Image questions (Fig. \ref{fig:prompt-ti}), 2) Image-Text questions (Fig. \ref{fig:prompt-it}), and 3) Image-Image questions (Fig. \ref{fig:prompt-ii}). We use GPT-4o (gpt-4o-2024-05-13) for question generation.

\lstset{
    basicstyle={\small},
    frame=single,
    breaklines=true,
}

\begin{figure*}[!h]
\begin{lstlisting}
Caption:
{textual_context}

You are provided with an image caption and a corresponding image. The question should be generated based on the **caption**, but the answer must come from the **visual or textual aspects of the image**.

### Important Instructions:
- The question must be derived from the caption, but the answer must depend on visual or textual details found in the image (e.g., objects, people, background, colors, clothing, etc.).
- Do not generate a question that can be answered directly from the caption. The answer should require information from the image.
- Do not generate a question based on the text found inside the image (for example, text from a poster or sign), but the answer can lie in the text inside image.
- The question should not give away the image directly, but it should allow the caption to guide the user to find the relevant image.
- Avoid speculative or ambiguous questions. Ensure the question and answer are logically connected to both the caption and the image.

Refer to the examples below for better understanding:
Example 1:
'question': "Is the girl, who is wearing a Flapper garb, wearing specs?", 
'choices': ['(A) No', '(B) Yes'], 
'answer': '(A)'
# Used caption: 'Photo of a girl in "flapper" garb. Taken in Moscow, Idaho in 1922. Donated by Dave Bumgardne., Flappers in New Woman Era, 1920s'

Example 2:
'question': "What is the color of the suit worn by Benedict?", 
'choices': ['(A) Black', '(B) Green', '(C) Grey', '(D) Brown'], 
'answer': '(C)'

# Used caption: 'The Way Way Back Australian Movie Premiere - Toni Collette At State Theatre, Sydney, Australia - 6th June 2013, Tori becomes involved in her first love triangle with Nate Cooper and Duncan Stewart (actor Benedict Wall pictured).'

After you generated the question, please also describe the caption you used as a fact to generate the question as follows:
"used textual facts": 'Polish football player Tomasz Frankowski Polski, Frankowski in 2010'

The output format should be:
"question" : generated question,
"choices" : generated options in a list,
"answer" : answer,
"used textual facts" : used caption
\end{lstlisting}
\caption{Prompt for Text-Image questions}
\label{fig:prompt-ti}
\end{figure*}

\begin{figure*}[!h]
\begin{lstlisting}
Caption:
{textual_context}

You are provided with images and their corresponding caption. The **question** should be generated based on the **visual aspects of the image** (such as people, objects, colors, background, etc.), and the **answer** must come from the **caption**.

### Important Instructions:
- The **question** must be based purely on the **visual aspects** of the image (e.g., objects, attire, setting, people, background, etc.).
- The **answer** must be derived only from the **caption** , and not from any visible text or visual aspects in the image.
- Ensure that the **question and answer** are logically connected to both the image and the caption.
- **Do not generate speculative or ambiguous questions**. Focus on visible aspects in the image and ensure that the answer is present in the caption.
- If the question has **textual overlap** with the image (e.g., a visible sign or poster), make sure that the answer comes from the **caption**, not from the visible text in the image.

Example 1:
'question': "The image shows a man with dark hair and a beard, dressed in a grey jacket, standing against a brightly colored background. Who is this person?", 
'choices': ['(A) Robbo', '(B) Benedict Wall', '(C) Jake Ryan', '(D) Ricky Sharpe'],
'answer': '(B)'
# The question is visually based, and the answer (Benedict Wall) comes from the **caption**.
Referred caption: 'The Way Way Back Australian Movie Premiere - Toni Collette At State Theatre, Sydney, Australia - 6th June 2013, Tori becomes involved in her first love triangle with Nate Cooper and Duncan Stewart (actor Benedict Wall pictured).'

Example 2:
'question': "The image shows a young woman wearing a short dress and a cloche hat. In which location was this photograph taken?", 
'choices': ['(A) New York City', '(B) Chicago', '(C) Moscow, Idaho', '(D) Los Angeles'],
'answer': '(C)'
# The question is visually based, and the answer (Moscow, Idaho) comes from the **caption**.
Referred caption: 'Photo of a girl in "flapper" garb. Taken in Moscow, Idaho in 1922.'

After you generated the question, please also describe the caption you used as a fact to get the answer as follows:
"used textual facts": 'Polish football player Tomasz Frankowski Polski, Frankowski in 2010'

The output format should be:
"question" : generated question,
"choices" : generated options in a list,
"answer" : answer,
"used textual facts" : used caption
\end{lstlisting}
\caption{Prompt for Image-Text questions}
\label{fig:prompt-it}
\end{figure*}

\begin{figure*}[!h]
\begin{lstlisting}
You are provided with an image. A question should be generated based purely on the **visual or textual aspects of the image** provided. Both the **question** and the **answer** should come from the **image itself**.

### Important Instructions:
- The **question** should describe a **visual or textual detail** present in the image.
- The **answer** must also be derived from the **visual elements** or **text** that is present within the image.
- Avoid using **external information** that is not visible in the image to form the question or answer.
- The image should be identifiable from the question.

Refer to the examples below for better understanding:
Example 1:
"question": "What color is the hat worn by the person standing against the white background with a cap?",
"choices": [ "(A) Red","(B) Blue", "(C) Green", "(D) Black" ],
"answer": "(A)"
 # both question and answer depend on the visual apsects of the image
Example 2:
"question": "What number is on the red jersey worn by the athlete in the outdoor stadium?",
"choices": [ "(A) 7", "(B) 10", "(C) 15", "(D) 3"],
"answer": "(C)"
# both question and answer depend on the visual apsects of the image

The output format should be:
"question" : generated question,
"choices" : generated options in a list,
"answer" : answer
\end{lstlisting}
\caption{Prompt for Image-Image questions}
\label{fig:prompt-ii}
\end{figure*}

\end{document}